\documentclass{article}

\usepackage{arxiv}

\usepackage[utf8]{inputenc} % allow utf-8 input
\usepackage[T1]{fontenc}    % use 8-bit T1 fonts
\usepackage{hyperref}       % hyperlinks
\usepackage{url}            % simple URL typesetting
\usepackage{booktabs}       % professional-quality tables
\usepackage{amsfonts}       % blackboard math symbols
\usepackage{nicefrac}       % compact symbols for 1/2, etc.
\usepackage{microtype}      % microtypography
\usepackage{lipsum}		% Can be removed after putting your text content
\usepackage{graphicx}
\usepackage{natbib}
\usepackage{doi}
\usepackage{svg}
\usepackage{amsmath}
\usepackage{caption}
\usepackage{subcaption}
\usepackage{float}

\title{Uncovering Bias in Face Generation Models}

\author{\href{1234-5678-9012}{\includegraphics[scale=0.06]{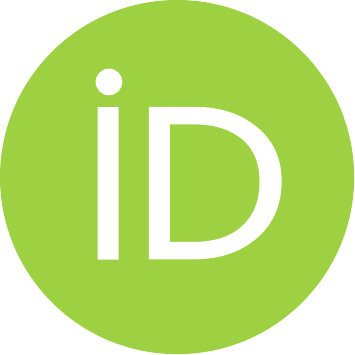}\hspace{1mm}Cristian Muñoz}\\
	Holistic AI\\
	\texttt{cristian.munoz@holisticai.com} \\
	\And
	\href{https://orcid.org/0000-0000-0000-0000}{\includegraphics[scale=0.06]{orcid.pdf}\hspace{1mm}Sara Zannone}\\
	Holistic AI\\
	\texttt{sara.zannone@holisticai.com} \\
	\And
	\href{https://orcid.org/0000-0000-0000-0000}{\includegraphics[scale=0.06]{orcid.pdf}\hspace{1mm}Umar Mohammed}\\
	Holistic AI\\
	\texttt{sara.zannone@holisticai.com} \\
	\And
	\href{https://orcid.org/0000-0000-0000-0000}{\includegraphics[scale=0.06]{orcid.pdf}\hspace{1mm}Adriano Koshiyama}\\
	Holistic AI\\
	\texttt{adriano.koshiyama@holisticai.com} \\
}

\renewcommand{\shorttitle}{Uncovering Bias in Face Generation Models}

\hypersetup{
pdftitle={A template for the arxiv style},
pdfsubject={q-bio.NC, q-bio.QM},
pdfauthor={David S.~Hippocampus, Elias D.~Striatum},
pdfkeywords={First keyword, Second keyword, More},
}

\begin{document}
\maketitle

\begin{abstract}
Recent advancements in GANs and diffusion models have enabled the creation of high-resolution, hyper-realistic images. However, these models may misrepresent certain social groups and present bias.  Understanding bias in these models remains an important research question, especially for tasks that support critical decision-making and could affect minorities. The contribution of this work is a novel analysis covering architectures and embedding spaces for fine-grained understanding of bias over three approaches: generators, attribute channel modifier, and post-processing bias mitigators. This work shows that generators suffer from bias across all social groups with attribute preferences such as between 75\%-85\% for whiteness and 60\%-80\% for the female gender (for all trained CelebA models) and low probabilities of generating children and older men. Modifier and mitigators work as post-processor and manipulate the attribute boundaries, these changes affect the generator performance. For instance, attribute channel perturbation strategies modify the embedding spaces. We quantify the influence of this change on group fairness by measuring the impact on image quality and group features. Specifically, we use the Fréchet Inception Distance (FID), the Face Matching Error and the Self-Similarity score. For Interfacegan, we analyze one and two attribute channel perturbations and examine the effect on the fairness distribution and the quality of the image. Finally, we analyzed the post-processing bias mitigators, which are the fastest and most computationally efficient way to mitigate bias. We find that these mitigation techniques show similar results on KL divergence and FID score, however, self-similarity scores show a different feature concentration on the new groups of the data distribution. The weaknesses and ongoing challenges described in this work must be considered in the pursuit of creating fair and unbiased face generation models. 
\end{abstract}

% keywords can be removed

\section{Introduction}

Generative Adversarial Networks (GANs) and Diffusion Models (DM) are relatively recent machine learning models that have been used, among other things, as generative models \cite{Goodfellow-et-al-2016}. These models have been used for a wide range of applications, such as image generation, style transfer, and video synthesis \cite{yang2022diffusion}\cite{lievrouw1994contemporary}\cite{kawar2022imagic}. One of the most recent and popular application of these models is face generation, which involves generating new realistic facial images. However, it is easy to see why implications around fairness and discrimination may arise in these cases. \\

It is a well-known fact that the datasets used in machine learning applications are not always representative of the whole population, and can in fact under-represent minorities \cite{pmlr-v81-buolamwini18a}\cite{wolfe2022markedness}. When training datasets are biased, the generated datsets will also be unbalanced. Since the generated datasets may be employed in a variety of applications, including as inputs to other AI systems, it is important to ensure that they display enough diversity, especially when it comes to protected attributes like gender, ethnicity or age. It is essential to perform a correct bias analyis of the generative models, and also select the best mitigation strategy for each specific case. \\

Measuring bias in GANs and DMs can be a complex task, as there is not a clear method or benchmark established yet, but is a crucial step to evaluate the effectiveness of the bias correction methods and to guarantee fairness in the generated images.  Here, we propose a variety of ways to measure the diversity of the output dataset, and investigate the mechanistic sources of bias in the networks.

\section{Related Works}

The most powerful deep learning models are usually trained by ingesting an incredible amount of information in an unsupervised fashion. Then,  strategies are created to extract and manipulate in an intentional manner the features learned by the model. Understanding the model behavior helps us create more efficient methods to control these large networks. This section covers the current most popular AI systems for Face Generation, the strategies used to control the face attributes, and mitigation techniques that improve the fairness of the model's outputs. Finally, we detail some previous works that analyse bias in face generation.
\subsection{Face Generators}
In the last few years, face generation models have shown remarkable performance on several tasks, particularly for high-resolution, hyperrealistic images, face generation \cite{karakas2022fairstyle} and face manipulation manipulation \cite{kammoun2022generative}. GAN-based and Diffusion-based models are the most popular strategies that use adverbial approach and variational inference during training.  Diffusion models present a high cost for both training and inference, however,  results suggest that these models are better at covering the data distribution \cite{rombach2021highresolution}.  
\subsubsection{GAN for Face Generation} GANs exploit adversarial training to build generators with excellent performance. There are numerous strategies that can be used to build GAN-based FG models, more details can be found \cite{kammoun2022generative}\cite{wang2022gan}\cite{ning2020multi}. In \cite{kammoun2022generative}, models are organized in three main types: Conditional, Controllable, and Progressive. Uncontrolled approaches have shown the best performance due to the big amount of unannotated data available. One of the most important families of GAN models is the StyleGAN family \cite{karras2019style}. StyleGAN2\cite{karras2020analyzing}, a SOTA model in facial generation which applies effective strategies to improve the model performance such as: generator normalization, revisiting progressive growing and regularizing the generator.  The model can be controlled modifying the input vector (Z and W) or some style layer parameters. CIPS \cite{anokhin2020image} is another GAN-Based model that uses random latent vectors and pixel coordinates to independently compute the color value at each pixel, eliminating the need for spatial convolutions or similar operations that propagate information across pixels. This new architecture is trained in an adversarial fashion and is said to produce similar image quality as SOTA methods.

\subsubsection{Diffusion Models for Face Generation (DM)} Diffusion models are a type of machine learning models that describe the evolution of a quantity over time, inspired by diffusion processes in physics. DMs define how information spreads and evolves over time, by modeling the change over a small time step as a combination of a drift term and a random fluctuation term. Denoising Diffusion Probabilistic Models (DDPM) \cite{ho2020denoising} is a DM that uses a weighted variational bound to achieve high-quality image synthesis through a connection with denoising score matching and Langevin dynamics. These models also offer a generalization strategy for autoregressive decoding. Diffusion Models have achieved state-of-the-art performance for image generation task on benchmark datasets such as CIFAR10\cite{krizhevsky2009learning} and LSUN\cite{song2015lsun}. Another interesting strategy that merges ideas from DMs and adversarial training is Latent Diffusion \cite{rombach2021highresolution}. Latent Diffusion Models (LDMs) apply a latent space of pre-trained autoencoders to produce high-quality and flexible results, while reducing computational requirements. They can achieve SOTA performance on image in-painting, unconditional image generation, semantic scene synthesis, and super-resolution, by introducing cross-attention layers into the model architecture.

\subsection{Face Attribute Controllers}

Attribute Controllers, also called mixed styles generation \cite{kammoun2022generative}, refer to the process of modifying multiple facial attributes simultaneously to generate faces in mixed styles using GANs, turning unconditional GANs into controllable GANs. This can be achieved through various methods, some of which will be cover below. These methods can be used to control a wide range of facial attributes, such as facial expressions, poses, hair styles, ages, and demographic characteristics such as race, gender and even emotions. A controller can modify the input vector space or space parameter to variate the face attributes. One of the best-known methods is Interfacega \cite{shen2020interfacegan}. This framework is used for semantic face editing by interpreting the latent semantics learned by GANs. Interfacegan finds that well-trained GANs have a disentangled representation after linear transformations. It allows for precise control of facial attributes, and it can be used for real image manipulation. Another space modifier is GANSpace\cite{härkönen2020ganspace}, this technique analyzes GANs and create interpretable controls for image synthesis by using PCA on `latent space`, and applying layer-wise perturbations along the principal directions. The method also allows for control of multiples GANs in a similar way as StyleGAN with good results. StyleSpace\cite{wu2021stylespace} analyzes the `latent style space` of StyleGAN2, which consist of channel-wise style parameters and is found to be more disentangled than previous strategies. This methodology enables the discovery of style channels that control specific visual attributes by using a pretrained classifier or a small number of example images.

One common problem with this approach is to disentangle attribute perturbation. For example, if we perturb the latent space that controls the gender attribute, we could also accidentally modify other undesired attributes, such as age or ethnic origin. These modifications are usually proportional to the scale of perturbation. InterfaceGAN and StyleGAN have guidelines for modifying the latent space or style latent to disconnect dependencies between facial attributes. The fact that there are alternatives that allow for the disconnection of these dependencies indicates the existence of multiple local optima that tend to demand of defining the boundary at which an attribute change. Therefore, if there are multiple solutions, what criterion allows me to define the optimal boundary for fairness in face generation?

\subsection{Bias in Machine Learning}
Generating synthetic images can help improve the accuracy of various classification tasks \cite{ xu2018fairgan}\cite{sattigeri2019fairness}, but it's also known that machine learning models can amplify biases present in the training data. This can lead to the creation of images with unintended biases. To address this issue, there has been a growing focus on analyzing\cite{wolfe2022markedness}, detecting and mitigating bias in deep learning models\cite{yu2020inclusive}\cite{choi2020fair}\cite{ mcduff2019characterizing}.  Bias mitigators con grouped in three categories: pre-processor, in-processor and post-processor. In this work we will focus on post-processor strategies.

\subsection{Post-processing Bias Mitigation for Face Generation}
Bias mitigation for face generation can lead to inaccuracies and unfair representations of certain groups of people. There are post-processing techniques that can help reduce these biases. One such technique is StyleFlow\cite{10.1145/3447648}, which focuses on this problem by formulating conditional exploration as a conditional continuous normalizing flow in the GAN latent space, conditioned by the attribute features. This method enables fine-grained disentangled edits along various attributes and has been evaluated using the face and car latent spaces of StyleGAN. Another well-known strategy is FairGen\cite{tan2020fairgen}, which utilizes GAN interpretation and a Gaussian Mixture Model (GMM) to produce images with a more balanced attribute distribution. By controlling the sampling of latent codes, FairGen has been shown to effectively reduce biases in generated images and can even be used to uncover biases in commercial face classifiers and super-resolution models. 
A different path to removing biases from generated images is to modify the model parameters themselves. This is where FairStyle \cite{karakas2022fairstyle} comes in - it directly modifies a pre-trained StyleGAN2 model to generate images with a more representative distribution of face attributes. The style space of the StyleGAN2 model is leveraged to control these attributes and reduce biases. It is important to note that bias mitigation for face generation is a complex issue and there's no one size fits all solution. It requires a combination of technical, ethical, and social considerations and ongoing research. Additionally, there's not a standard methodology or evaluation protocol for fairness in FG, and there are trade-offs between diversity and the realism of the generated images.

\section{Methodology}
In this section, we describe the methodology employed for the analysis of fairness in attribute controllers and bias mitigators. The analysis is conducted at three levels. Firstly, by examining the presence of bias in generator models. Secondly, by analyzing the presence of bias when perturbing the attribute boundaries defined by the controllers. And thirdly, by evaluating the effects on data generation caused by the mitigation strategy. This analysis allows us to gain a deeper understanding of the subject and identify potential areas for further research in this field.

\subsection{Attribute Classifier}

In order to evaluate the various characteristics of images generated by GANs, we employed the FaceNet model for face matching\cite{serengil2020lightface} and face attribute analysis  \cite{serengil2021lightface} from the Deepface library. This library is equipped with both facial recognition capabilities and the ability to analyze facial attributes such as age, gender, emotion, and ethnicity. The facial recognition component of the model performed exceptionally well in experiments using the Label Faces from the Wild (LFW) data, achieving an accuracy rate of 99.2\%. For more information about the model, please refer to the following link \footnote{https://github.com/serengil/deepface}. The model is also capable of classifying facial variations, including gender (female, male), ethnicity (white, Latino/Hispanic, black, Asian, Indian, Middle Eastern), and emotion (happy, angry, neutral, surprised, fearful, sad, disgusted). To evalute the interfacegan with 2 channel perturbation and bias mitigators we use classifier\footnote{https://catlab-team.github.io/fairstyle/} which was used for the FairStyle \cite{karakas2022fairstyle} strategy. 

\section{Results}
In this work, we conduct an in-depth examination of bias in face generation models. First, we examine whether the generated dataset is balanced for people of different gender, race and age. We also include an analysis of the emotions displayed by different subjects. Then, we alter the latent space of the network models to get further insight into the mechanistic causes of bias. To accomplish this, we analyse various attribute controllers and bias mitigation techniques. Through this investigation, we are able to gain a deeper understanding of the crucial considerations that must be taken into account for improving the effectiveness of the bias correction.

\subsection{Bias in Face Generation Models}
\begin{figure}[H]
  \centering
    \subfloat[a]{
        \includegraphics[width=\textwidth]{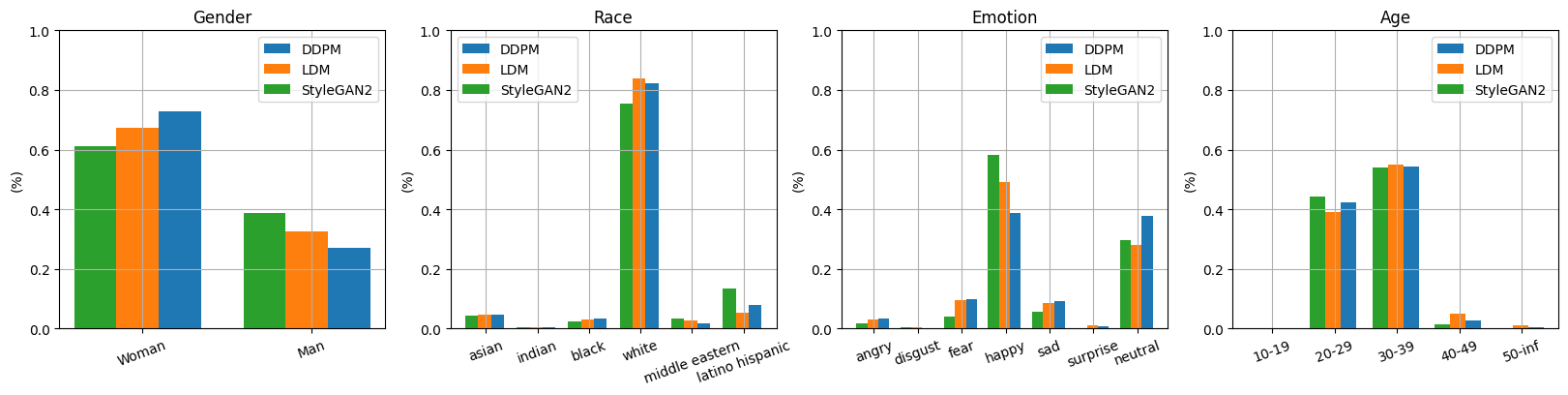}
        \label{fig:a}%
    }
    \hfill%
    \subfloat[a]{
        \includegraphics[width=\textwidth]{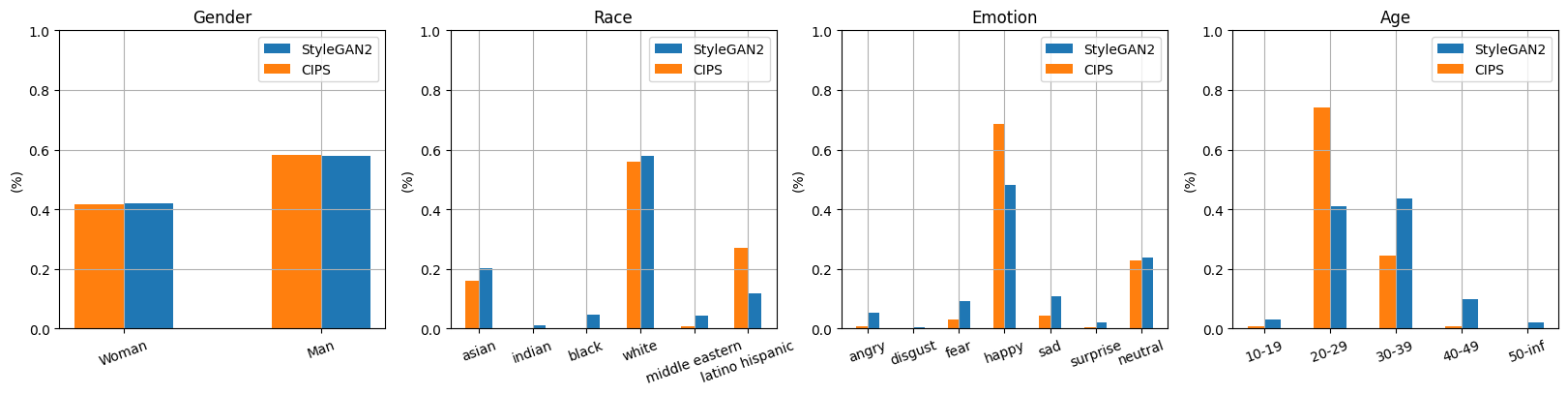}
        \label{fig:b}%
    }
  \caption{Face Attribute Distribution of models for dataset (a) CelebA dataset and (b) FFHQ dataset.}\label{fgmodel:bias}
  %\Description{A woman and a girl in white dresses sit in an open car.}
\end{figure}

If the datasets used to train face generation models are not a representative sample of the population, the output images will also contain a similar bias. As a first step, we analyse how the generated faces are distributed when it comes to gender, ethnicity, age and emotion. For our analysis, we aggregated the results according to the training dataset. The StyleGAN2-FFHQ and CIPS models were trained on the FFHQ dataset, while StyleGAN2-CelebA, LDM and DDPM models were trained on the CelebA dataset. Figure \ref{fgmodel:bias} (a) and (b) displays the results of our analysis for the models trained on the CelebA and FFHQ dataset, respectively. The generated datasets seem to display bias for virtually every considered attribute. \\

% \begin{itemize}
% \item 
Both sets of models show some gender bias. However, models trained on the FFHQ dataset demonstrate a preference for male face generation, while models trained on the CelebA dataset prefer female faces. In the case of CelebA, the preference is more pronounced, reaching over 70\% for the DDPM model. 
%\item 
The disparity seems to be even more striking when we examine ethnicity: all models show a strong preference for the whiteness category, reaching, in the case of CelebA, values that exceed 80\%. This means that less than $20\%$ of the generated dataset will belong to an ethnicity other than white.  
%\item 
The output dataset is unbalanced when it comes to age too. All models generate mostly faces categorized in the age range of 20 to 40 years. The CIPS model shows stronger preference for the 20-29 age group. 
%\item 
When it comes to emotions, it appears that most of the images generated display people looking happy, or neutral, with little variety when it comes to other, comparatively negative, types of emotion.  \\
%\end{itemize}

Overall, the output datasets seem to uncover the presence of some serious bias, especially when it comes to ethnicity. 
%Although the emotion attribute is not considered a social analysis group, it is important to guarantee some diversity in this sense as well. If we use our generated dataset as an input for another AI system, such as a face matching model, we need to ensure that the algorithm is tested across a wide range of emotional states. Facial expression can can indeed sometimes affect the performance of those AI systems that use the generated images. bias present in this attribute can affect the AI systems that use the images generated by boosting itself. In addition to bias, there are other important factors that models have to deal with: diversity and quality.

\subsubsection{Representative Similarity in Face Generation Models}
The Self-Similarity score quantifies the degree to which face features are shared within a group. A high value indicates that the images share many characteristics. This value could be used as an indicator to compare the diversity between groups. It should be considered that this is only a relative value which should be examined together with other metrics, since distortion or poor image quality can lead to false positives under this approach. \\

In Figure \ref{fgmodel:selfsim_facattr}, we observe no obvious relationship between diversity and bias. The CIPS, DDPM and StyleGAN-CelebA models all display a generally high score, which would indicate many similar features in the generated images. On the other hand, the StyleGAN2 and LDM model tend to score in the lower values for gender, race, and emotion. This suggests that the output images are relatively diverse for all of these attributes. However, every single one of the models examined seems to display a high self-simillarity score for the age attribute. What could cause this lack of diversity when it comes to age? \\

A key factor is the diversity of examples. In the Figure \ref{fgmodel:gender_vs_age} we observe the results of our intersectional analysis for gender and age. We can see, for example, that the LDM model was not able to generate images of males in the 10-19 age range (you can review Appendix 1.3). All images that were generated in this age region were categorized as women. Indeed, equality at the intersection of multiple face attributes is one of the problems that most hinders mitigation strategies. \\

\begin{figure}[htp] 
    \centering
    \subfloat[a]{%
        \includegraphics[width=0.47\textwidth]{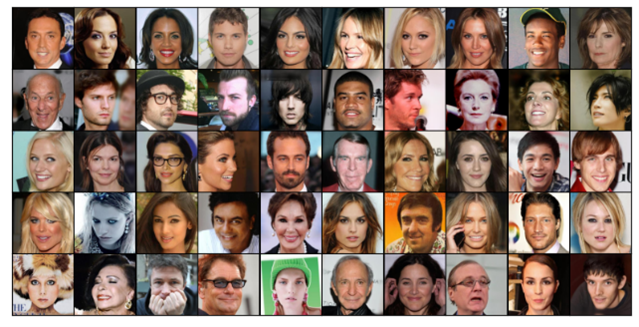}%
        \label{fig:a}%
        }%
    \hfill%
    \subfloat[b]{%
        \includegraphics[width=0.47\textwidth]{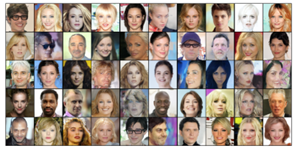}%
        \label{fig:b}%
        }%
    \hfill%
    \subfloat[c]{%
        \includegraphics[width=0.47\textwidth]{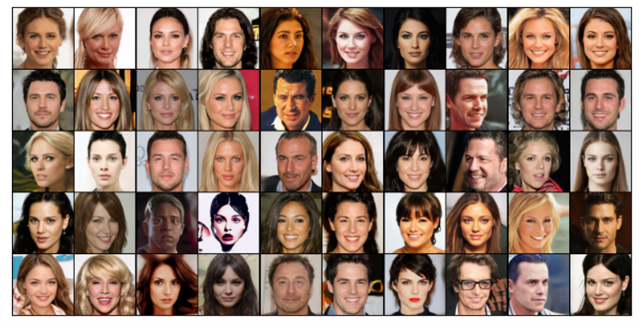}%
        \label{fig:c}%
        }%
    \hfill%
    \subfloat[d]{%
        \includegraphics[width=0.47\textwidth]{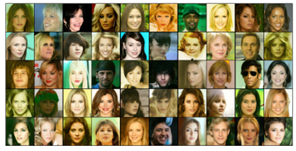}%
        \label{fig:d}%
        }%
    \caption{Image examples from: (a) CelebA dataset. (b) Latent Diffusion Model. (c) StyleGAN2. (d) Denoising Diffusion Probabilistic model.}\label{ctrl:img_gen}
\end{figure}

Next, we performed a global analysis of the generated data, Table \ref{fgmodel:statistics} presents the following image quality metrics: FID Score, Face Matching Error, and Global Self-Similarity. We considered both the CelebA and FFHQ training datasets. Interestingly, the StyleGAN models present the lowest self-similarity score and the highest image quality (lowest FID score), when trained on the FFHQ dataset. However, for the CelebA dataset, it is DDPM that displays the best image quality (lowest FID Score). Conducting a review of the generated datasets, it was observed that even with a low FID value, LDM creates many images with faces easily identifiable as fake. We show some example images from the CelebA dataset, and from the datasets generated by the Latent Diffusion Model, StyleGAN2 and Denoising Diffusion Probabilitic Model (Figure \ref{ctrl:img_gen} (a)(b)(c)(d) respectively). %These examples can increase the number of non-shared characteristics and cause a reduction in self-similarity, but they also have the highest percentage of images that make face detection difficult.

\begin{table}[h]
\centering
\caption{Model Statistics}\label{fgmodel:statistics}
\begin{tabular}{cccccc}
\hline
Model     & Dataset & Image Size & FID   & \%Error FM & Self-Similarity \\ \hline
DDPM      & CelebA  & 256x256    & 22.66 & 4.48\%     & 0.378           \\
LDM       & CelebA  & 256x256    & 27.14 & 5.38\%     & 0.057           \\
StyleGAN2 & CelebA  & 256x256    & 30.28 & 1.16\%     & 0.354           \\ \hline
StyleGAN2 & FFHQ    & 1024x1024  & 22.54 & 11.22\%    & 0.061           \\
CIPS      & FFHQ    & 1024x1024  & 68.39 & 0.22\%     & 0.267           \\ \hline \cline{3-6} 
\end{tabular}
\end{table}

%An interesting case is the models trained with the CelebA dataset. The DDPM model presents the best FID

%Mention that for example even if the DDPM model has a low FID value in self-similarity, then the DDPM has good image qualities but has little diversity.

\subsection{Face Attribute Controller}
For this analysis, we used controllers in various vector spaces to alter the preference of the attributes. For interfacegan, this is observed in the Z and W spaces. For GANSPace, we chose component 2 and 13, which refer to Gender and Age respectively. This selection was carried out manually, identifying the best components to control these attributes. For interfacegan and GANSpace controllers, the perturbation domain was between -2 and 2, while it was between -30 and 30  for StyleSpace. All controllers modify the StyleGAN2 model trained on the FFHQ dataset. For each perturbation, we generated 10000 output images, which were then analyzed and used to produced the statistics presented here. Figure \ref{ctrl:gender_attr} and Figure \ref{ctrl:age_attr} show the effect of perturbing the control channel of an attribute using different strategies. We can notice that changing one attribute often affects all the other attributes. This implies that the attributes are correlated in the model. For example, positively scaling the gender attribute reduces the preference for the happy characteristic for both interfacegan (Z and W) and stylespace. When the same perturbation is applied to GANSpace, the happiness feature increases slightly. Similarly, negatively perturbing  the gender attribute increasaes the preference for the white category for StyleGAN (Z), GANSpace and StyleSpace, but for GANSpace this preference decreases.

Interestingly, we can see in Figure \ref{ctrl:gender_attr} that perturbing the Age attribute for stylespace, causes more complex changes in the race, gender and emotion attributes. This gives us an idea of why it is difficult to mitigate bias in this attribute.

\subsubsection{Representational Similarity, Face Match Error and FID Score}
\begin{figure}[htp] 
    \centering
    \subfloat[a]{%
        \includegraphics[width=0.30\textwidth]{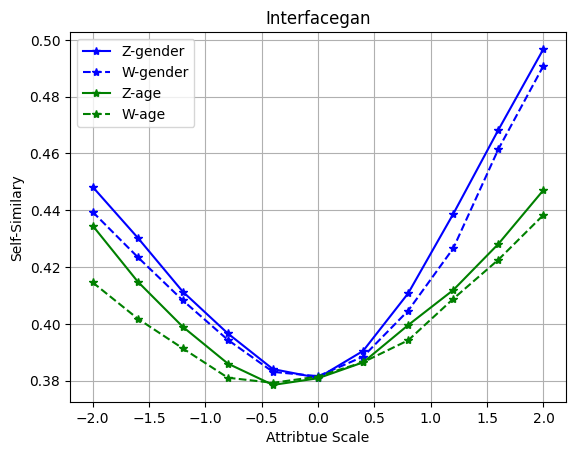}%
        \label{fig:a}%
        }%
    \hfill%
    \subfloat[b]{%
        \includegraphics[width=0.30\textwidth]{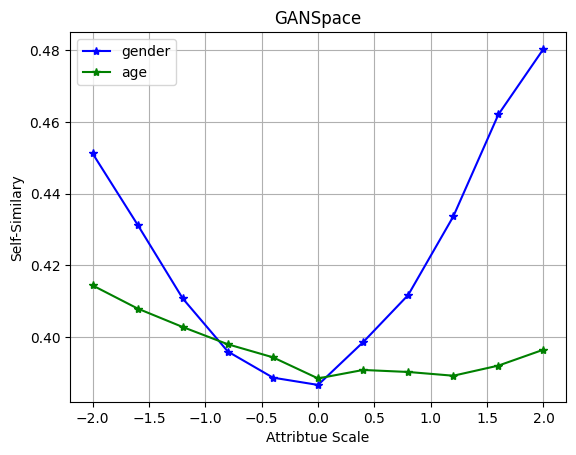}%
        \label{fig:b}%
        }%
    \hfill%
    \subfloat[c]{%
        \includegraphics[width=0.30\textwidth]{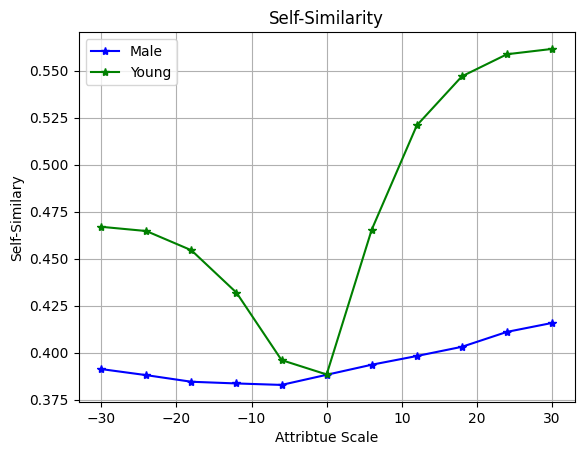}%
        \label{fig:c}%
        }%
    \hfill%    
    \subfloat[d]{%
        \includegraphics[width=0.30\textwidth]{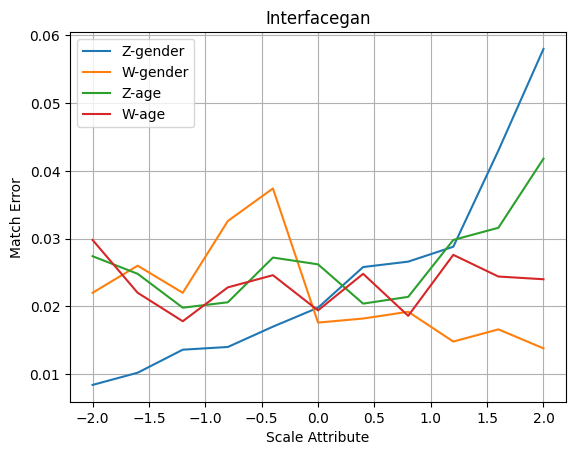}%
        \label{fig:a}%
        }%
    \hfill%
    \subfloat[e]{%
        \includegraphics[width=0.30\textwidth]{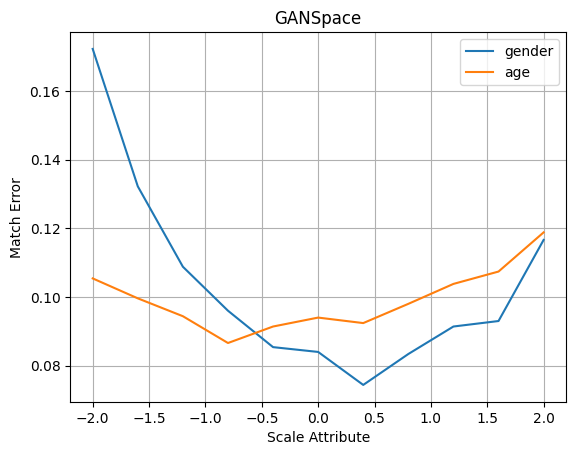}%
        \label{fig:b}%
        }%
    \hfill%
    \subfloat[f]{%
        \includegraphics[width=0.30\textwidth]{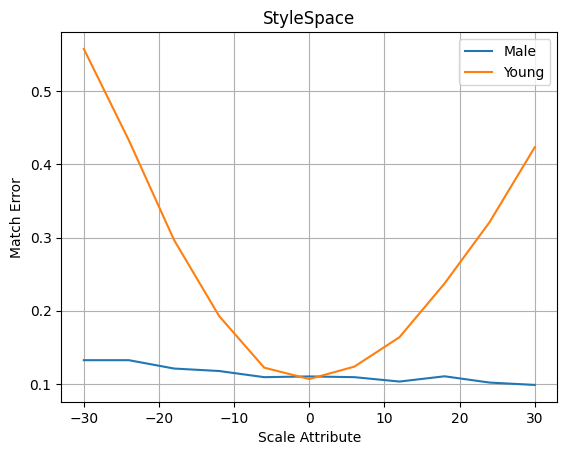}%
        \label{fig:c}%
        }%
    \hfill%    
    \subfloat[g]{%
        \includegraphics[width=0.30\textwidth]{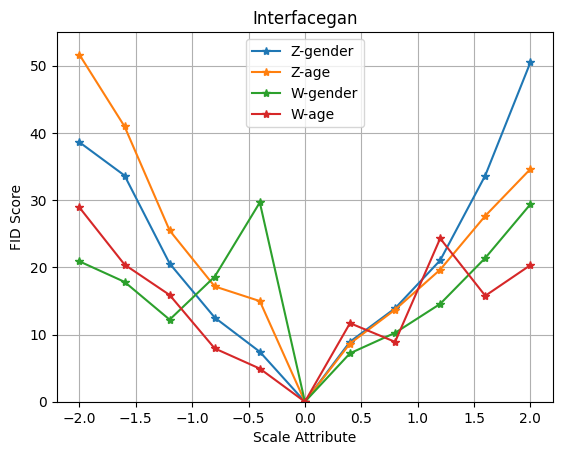}%
        \label{fig:a}%
        }%
    \hfill%
    \subfloat[h]{%
        \includegraphics[width=0.30\textwidth]{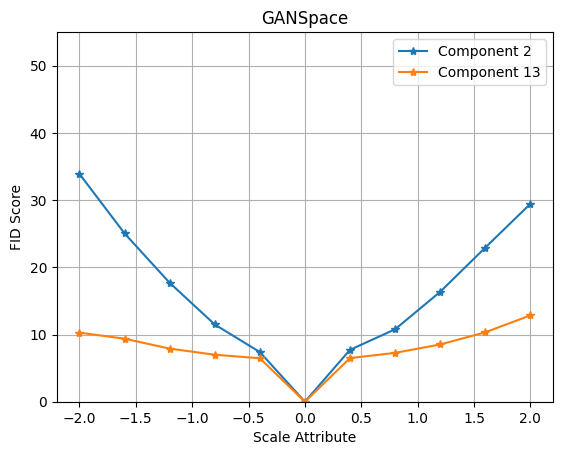}%
        \label{fig:b}%
        }%
    \hfill%
    \subfloat[i]{%
        \includegraphics[width=0.30\textwidth]{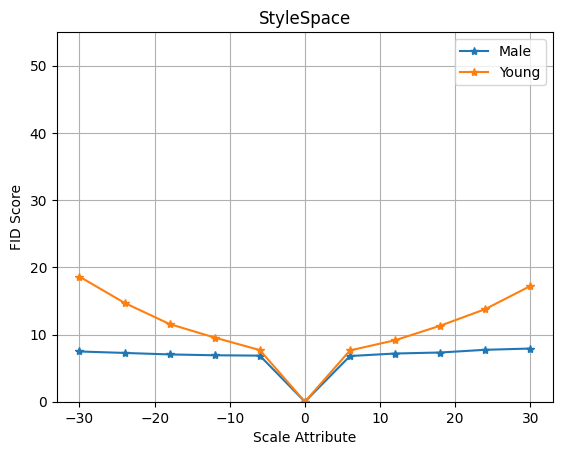}%
        \label{fig:c}%
        }%
    \caption{Performance metrics: The rows show the behaviour Self-Similaty, Face Matching error, FID Score. Each columns represent the Attribute Controller: Interfacegan, GANSpace and Stylespace.}\label{ctrl:metrics_plot}
\end{figure}

\begin{figure}[htp]
  \centering
  \includegraphics[width=0.8\linewidth]{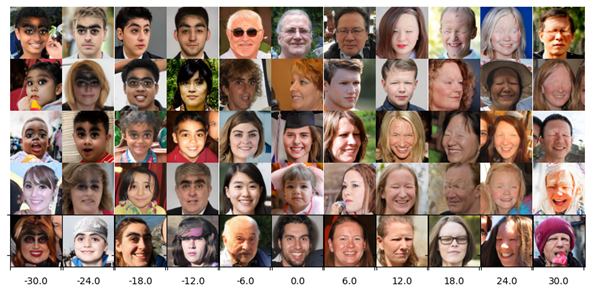}
  \caption{Image samples perturbing Young channel attribute from StyleSpace. The scale used was from -30 to 30. Can be oserved face distortion outside the interval of -6 to 6.}\label{ctrl:stylespace_age_perturbation}
  %\Description{A woman and a girl in white dresses sit in an open car.}
\end{figure}

For each of the perturbations introduced in the previous section, we also investigated the impact on the new data distribution. Figure \ref{ctrl:metrics_plot} (a) and (b) show that perturbing the gender attribute quickly increases the self-similarity score for the interfacegan and GANSpace controllers. This becomes more evident when comparing these results to the stylespace gender attribute curve in Figure \ref{ctrl:metrics_plot} (c), which, when negatively perturbed, manages to reduce the self-similarity score even further. If we compare it to the Match Error results in Figure \ref{ctrl:metrics_plot} (d),(e) and (f), we can see that the error curve for the stylespace age attribute controller is much higher compared to the other controllers. To better understand the reason behind this increase in error, Figure \ref{ctrl:stylespace_age_perturbation} shows a list of 5 images extracted for each perturbation that was performed using stylespace. It can be seen that for values outside the [-6 , 6] domain, the controller performs changes in the eyes that distort the generated image. This distortion cannot be spotted using the FID metric (Figure \ref{ctrl:metrics_plot} (i)). However, we can notice an anomaly in the self-similarity score ( Figure \ref{ctrl:metrics_plot} (c)) and Match Error (Figure \ref{ctrl:metrics_plot} (f)). This is because the eye distortion is a characteristic that all generated images will tend to share, and this would increase the similarity of the generated images.

The FID score could not detect this image distortion, but due to the detection error and the self-similarity score, we could note an anomaly in the generated data. This indicates that calculating one metric may not always be sufficient. Instead, our analysis suggests that a variety of quality and diversity metrics should be calculated in order to correctly evaluate the models. 

\subsubsection{Disambiguation of face attributes}
\begin{figure}[H]
  \centering
  \includegraphics[width=\linewidth]{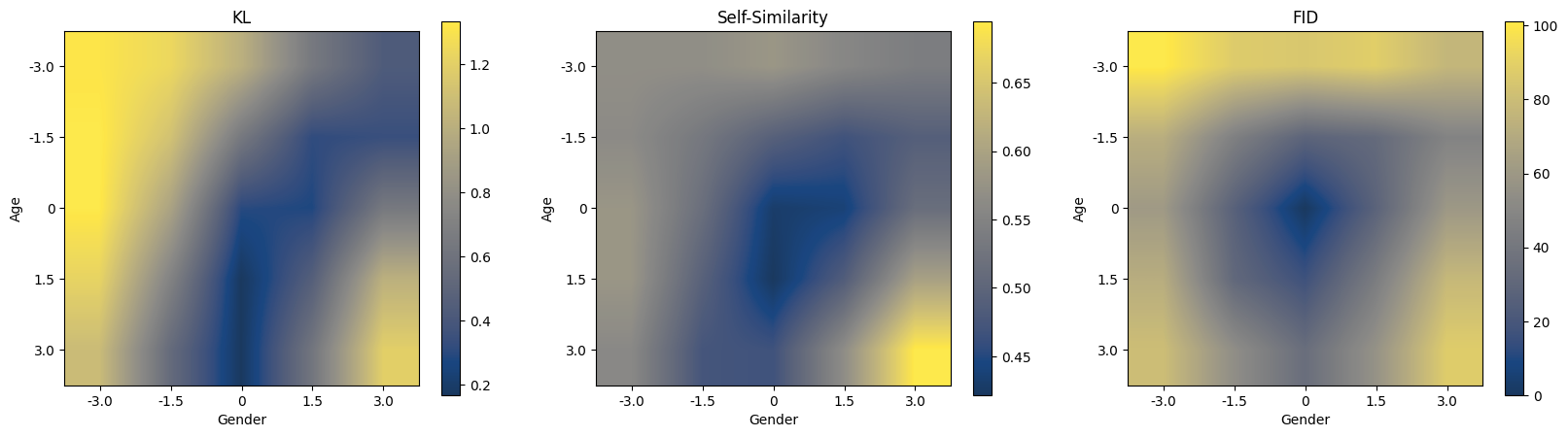}
  \caption{KL, Self-Similarity and FID score metrics in 2 attribute perturbation: gender and age. The age attribute used is the age boundary conditioned with gender attribute.}\label{bm:performance_metrics_2d}
  %\Description{A woman and a girl in white dresses sit in an open car.}
\end{figure}

Interfacegan proposes an alternative to disambiguating disturbance in attribute channels using conditional boundaries. This alternative allows us to explore the effect of perturbing more than one attribute space. For this, we used the boundary for gender alone, and the boundary for gender conditioned on the age attribute. The conditional boundary keeps the age vector from modifying the gender attribute, or at least it reduces its effect. The perturbation on the gender channel acts from [-3,3]. For this experiment, the CelebFaces\footnote{https://catlab-team.github.io/fairstyle/} classifier was used, following \cite{karakas2022fairstyle}, which has the binary category Young/Adult and Male/Female Interfacegan. The optimal distribution for group equity: [Young-Man, Adult-Man,Young-Woman, Adult-Woman] is [1/4, 1/4, 1/4, 1/4]. For each perturbation on the attribute channel, the distribution of categories was analyzed using 3 performance metrics: FID score, Self-Similarity and the KL divergence, which calculate how close distribution of categories is to a uniform distribution. Our results show that for a small neighborhood the FID score and Self-Similarity increase rapidly, while the increment is smaller in the direction where KL decreases (Figure \ref{bm:performance_metrics_2d}). The image was interpolated using Bi-linear sampling: a detailed plot using a 5x5 matrix is showed in Figure \ref{inter:2channel}. This related idea could help finding more effective and efficient strategies to mitigate bias.

From the top exposed image the best setting is a scale of Age=3 and a scale of Gender=0. In this case the attribute distribution was: [0.045, 0.318, 0.389, 0.247].

\subsection{Bias Mitigation}
\begin{figure}[H] 
    \centering
    \subfloat[a]{
        \includegraphics[width=0.4\textwidth]{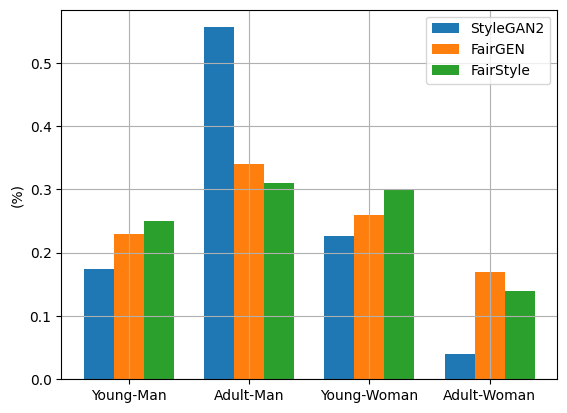}%
        \label{fig:a}%
        }%
    \hfill%
    \subfloat[b]{%
        \includegraphics[width=0.4\textwidth]{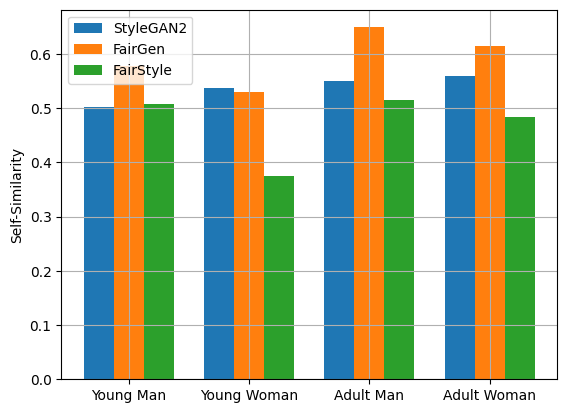}%
        \label{fig:b}%
        }%
    \caption{Bias Performance Metrics for Mitigators 2 fairness group Young-Gender: (a) Data Distribution. (b) Self-Similarity.}\label{bm:data_distribution}
\end{figure}

Mitigation strategies, in most cases, manage to adjust the data distribution to a proportion of unbiased attributes. However, for the case of Gender and Young, the mitigation still does not achieve an optimal distribution of attributes. In Figure \ref{bm:data_distribution}, the results are presented by replicating the FairStyle experiments and using the FairGen trained model. The KL results are very similar to those described in the literature \cite{karakas2022fairstyle}. An important analysis is to understand how the self-similarity value changes after the mitigation. In Figure \ref{bm:performance_metrics} it can be observed that, even if the overall value changes a little, the values by fairness group save important information about the new distributions. In Figure \ref{bm:data_distribution} (b) it can be seen that the two algorithms decrease the score for the “Young Woman” group, but the FairStyle algorithm also increases the “Young Man” and decreases the other categories. This is probably due to the reduction of very similar examples leaving the images with different invoices. Unlike the FairGen model, apparently the groups generated by the algorithm have a lot of concentration and therefore more similar images per group.
Error detection is also an anomaly detection strategy, but it has a high cost since you have to run a detector model to perform the evaluation.

\begin{table}[H]
\centering
\caption{Bias Performance Metrics for mitigators.}\label{bm:performance_metrics}
\begin{tabular}{cccc}
\hline
Metrics         & StyleGAN2 & FairGEN  & FairStyle \\ \hline
Self-Similarity & 0.377     & 0.364    & 0.402     \\
FID             & -         & 26.267   & 24.737    \\
KL              & -         & 3.00E-02 & 2.08E-02  \\ \hline
\end{tabular}
\end{table}

\section{Conclusion}
Face generation models have managed to recreate super-realistic images, but the bias problem involves more criteria than just the proportion of generated images. Models show a much greater preference for groups such as White (Race), and age attributes between 20 to 40 years. Models trained with the FFHQ dataset have a preference for male characteristics and models trained with CelebA have a preference for females. Most models fail to generate female gender attributes outside the 20-40 age ranges. Image semantic quality and feature diversity persistence in interest groups are necessary to improve our models. We also observed that very large perturbations in the style space add unwanted features that distort the image. These distortions are not necessarily detected by the FID score, so it is important to accompany the evaluations with additional metrics such as self-similarity, or the use of face detectors to detect anomalies in the newly fitted model.

%\section{Acknowledgments}
%Holistic AI Research Team.

\section{Appendices}

%\begin{verbatim}
\appendix

\section{Face Generation Model}
% \usepackage{multirow}
\subsection{Distribution of face attributes of generation models}

\begin{table}[H]
\caption{Data distribution for Face Attribute on each Generator Model.}
\begin{tabular}{ccccccc}
\hline
Attribute & Category & \multicolumn{5}{c}{Models}                                               \\ \cline{3-7} 
                           &                           & DDPM-CelebA & LDM-CelebA & StyleGAN2-CelebA & StyleGAN2-FFHQ & CIPS-FFHQ \\ \hline
Gender    & Woman                     & 0.4397      & 0.1035     & 0.4582           & 0.1015         & 0.3081    \\
                           & Man                       & 0.4054      & 0.0434     & 0.4708           & 0.0575         & 0.3183    \\ \hline
Race      & asian                     & 0.4582      & 0.1261     & 0.4194           & 0.2255         & 0.3869    \\
                           & indian                    & 0.3846      & 0.0747     & 0.4150           & 0.1395         & 0.2004    \\
                           & black                     & 0.5184      & 0.1914     & 0.5447           & 0.2166         & 0.3664    \\
                           & white                     & 0.4021      & 0.0711     & 0.3777           & 0.0691         & 0.2742    \\
                           & middle eastern            & 0.4126      & 0.0853     & 0.4770           & 0.0987         & 0.2247    \\
                           & latino hispanic           & 0.3865      & 0.0843     & 0.3789           & 0.1129         & 0.3402    \\ \hline
Emotion   & angry                     & 0.3891      & 0.0653     & 0.3789           & 0.0541         & 0.2239    \\
                           & disgust                   & 0.3662      & 0.0987     & 0.4712           & 0.1067         & ---       \\
                           & fear                      & 0.4433      & 0.0710     & 0.4311           & 0.0705         & 0.2132    \\
                           & happy                     & 0.3835      & 0.0697     & 0.3906           & 0.0678         & 0.2814    \\
                           & sad                       & 0.4165      & 0.0628     & 0.3688           & 0.0622         & 0.2565    \\
                           & surprise                  & 0.4222      & 0.0614     & 0.3432           & 0.1192         & 0.2078    \\
                           & neutral                   & 0.3988      & 0.0613     & 0.3727           & 0.0685         & 0.2822    \\ \hline
Age       & 10-19                     & 0.5386      & 0.4829     & 0.7442           & 0.4383         & 0.5461    \\
                           & 20-29                     & 0.3911      & 0.0627     & 0.4025           & 0.0814         & 0.3001    \\
                           & 30-39                     & 0.3993      & 0.0727     & 0.4030           & 0.0627         & 0.2172    \\
                           & 40-49                     & 0.4588      & 0.0987     & 0.4809           & 0.0870         & 0.2943    \\
                           & 50-inf                    & 0.5685      & 0.1813     & 0.6338           & 0.1197         & 0.3502    \\ \hline
\end{tabular}
\end{table}

\subsection{Self-Similarity of generation models by face attribute.}
% \usepackage{multirow}
\begin{table}[H]
\caption{Self-Similarity of Generator Model for each Face Attribute group}
\begin{tabular}{ccccccc}
\hline
Attribute & Category & \multicolumn{5}{c}{Models}                                               \\ \cline{3-7} 
                           &                           & DDPM-CelebA & LDM-CelebA & StyleGAN2-CelebA & StyleGAN2-FFHQ & CIPS-FFHQ \\ \hline
Gender    & Woman                     & 0.43974     & 0.10348    & 0.45822          & 0.1015         & 0.30807   \\
                           & Man                       & 0.40538     & 0.04343    & 0.47082          & 0.05753        & 0.31827   \\ \hline
Race      & asian                     & 0.45825     & 0.1261     & 0.41942          & 0.22546        & 0.38692   \\
                           & indian                    & 0.38458     & 0.07469    & 0.41502          & 0.13946        & 0.20037   \\
                           & black                     & 0.51843     & 0.19144    & 0.54473          & 0.21656        & 0.3664    \\
                           & white                     & 0.40211     & 0.07112    & 0.37773          & 0.0691         & 0.27415   \\
                           & middle eastern            & 0.41255     & 0.08532    & 0.47704          & 0.09866        & 0.22468   \\
                           & latino hispanic           & 0.38651     & 0.0843     & 0.37886          & 0.11288        & 0.34016   \\ \hline
Emotion   & angry                     & 0.38908     & 0.06532    & 0.37888          & 0.05412        & 0.22391   \\
                           & disgust                   & 0.3662      & 0.09867    & 0.47117          & 0.10672        & ---       \\
                           & fear                      & 0.44335     & 0.07102    & 0.43112          & 0.07046        & 0.21316   \\
                           & happy                     & 0.38351     & 0.06974    & 0.39061          & 0.06778        & 0.28135   \\
                           & sad                       & 0.4165      & 0.06284    & 0.36884          & 0.06224        & 0.25652   \\
                           & surprise                  & 0.42216     & 0.06138    & 0.3432           & 0.11917        & 0.20775   \\
                           & neutral                   & 0.39875     & 0.06131    & 0.37274          & 0.06851        & 0.28215   \\ \hline
Age       & 10-19                     & 0.53858     & 0.48292    & 0.74418          & 0.4383         & 0.54614   \\
                           & 20-29                     & 0.39113     & 0.06271    & 0.40254          & 0.08135        & 0.30008   \\
                           & 30-39                     & 0.39931     & 0.07266    & 0.40302          & 0.06266        & 0.21719   \\
                           & 40-49                     & 0.45882     & 0.09872    & 0.48089          & 0.08697        & 0.29432   \\
                           & 50-inf                    & 0.56848     & 0.18127    & 0.63376          & 0.11967        & 0.35018   \\ \hline
\end{tabular}
\end{table}

\subsection{Data Distribution grouped by gender and age attribute.}
% \usepackage{multirow}
\begin{table}[H]
\caption{Data Distribution grouped by Gender and Age Range}
\begin{tabular}{cccccccccccc}
\hline
Modelo & Dataset & \multicolumn{2}{c}{\textless{},19{]}} & \multicolumn{2}{c}{20-29} & \multicolumn{2}{c}{30-39} & \multicolumn{2}{c}{40-49} & \multicolumn{2}{c}{{[}50,\textgreater{}} \\ \cline{3-12} 
                        &                          & Man               & Woman             & Man         & Woman       & Man         & Woman       & Man         & Woman       & Man                 & Woman              \\ \hline
DDPM                    & CelebA                   & 0.001             & 0.000             & 0.152       & 0.261       & 0.094       & 0.433       & 0.013       & 0.012       & 0.003               & 0.002              \\
LDM                     & CelebA                   & 0.001             & 0.000             & 0.162       & 0.223       & 0.124       & 0.416       & 0.027       & 0.021       & 0.007               & 0.004              \\
StyleGAN2               & CelebA                   & 0.000             & 0.000             & 0.290       & 0.152       & 0.084       & 0.456       & 0.012       & 0.003       & 0.002               & 0.000              \\
CIPS                    & FFHQ                     & 0.004             & 0.003             & 0.452       & 0.288       & 0.118       & 0.126       & 0.007       & 0.001       & 0.001               & 0.000              \\
StyleGAN2               & FFHQ                     & 0.017             & 0.015             & 0.253       & 0.158       & 0.215       & 0.222       & 0.077       & 0.023       & 0.017               & 0.003              \\ \hline
\end{tabular}
\end{table}

\subsection{Model Self-Similarity}

\begin{figure}[H]
  \centering
  \includegraphics[width=\linewidth]{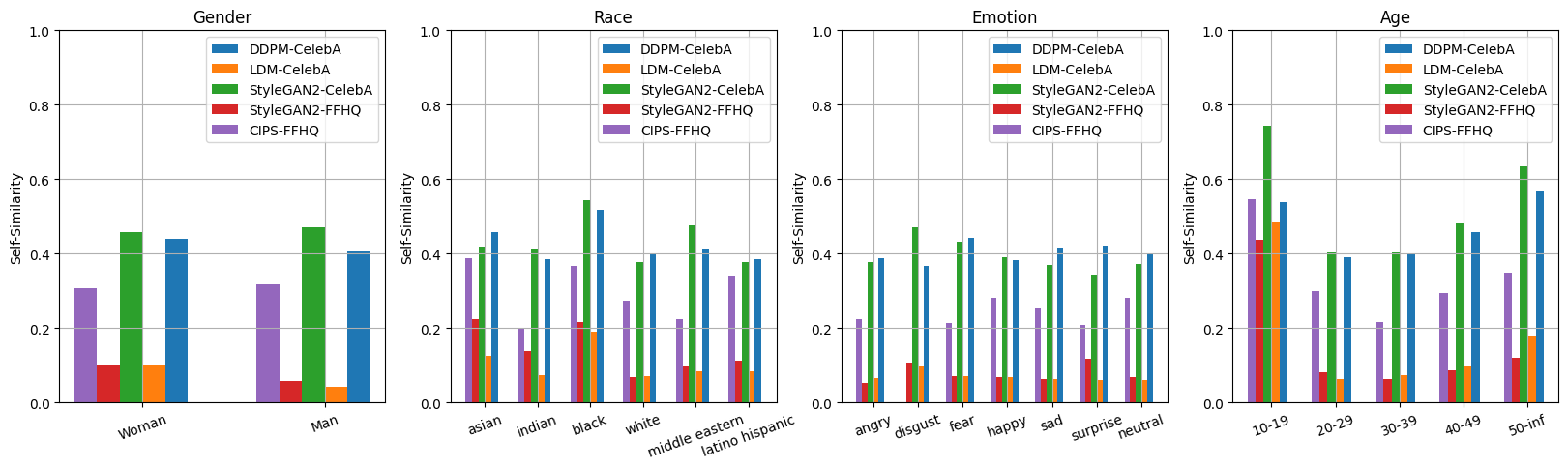}
  \caption{Model Self-Similarity by face attribute.}\label{fgmodel:selfsim_facattr}
\end{figure}

\begin{figure}[H]
  \centering
  \includegraphics[width=\linewidth]{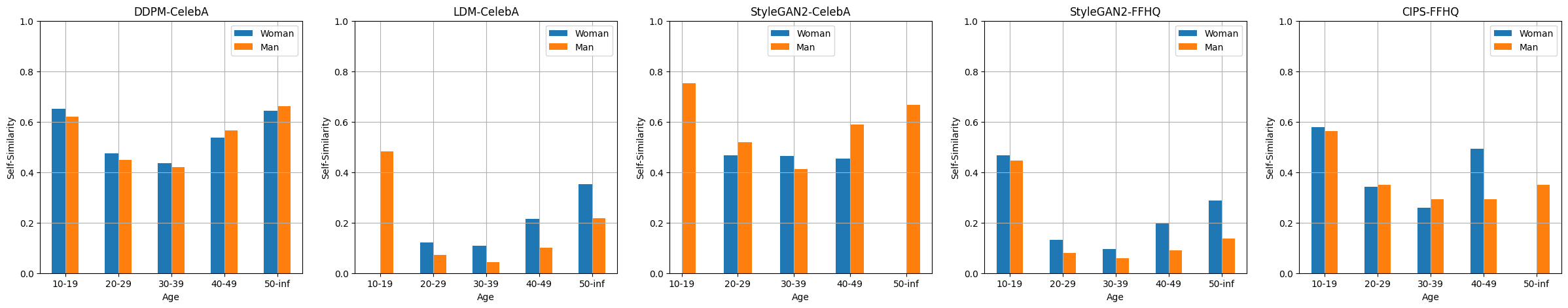}
  \caption{Model Self-Similarity grouped by gender and age.}\label{fgmodel:gender_vs_age}
\end{figure}

\section{Attribute Controllers}
\subsection{One channel attribute perturbation for different embedding spaces.}

\begin{figure}[H] 
    \centering
     \begin{subfigure}[b]{0.9\textwidth}
        \includegraphics[width=\textwidth]{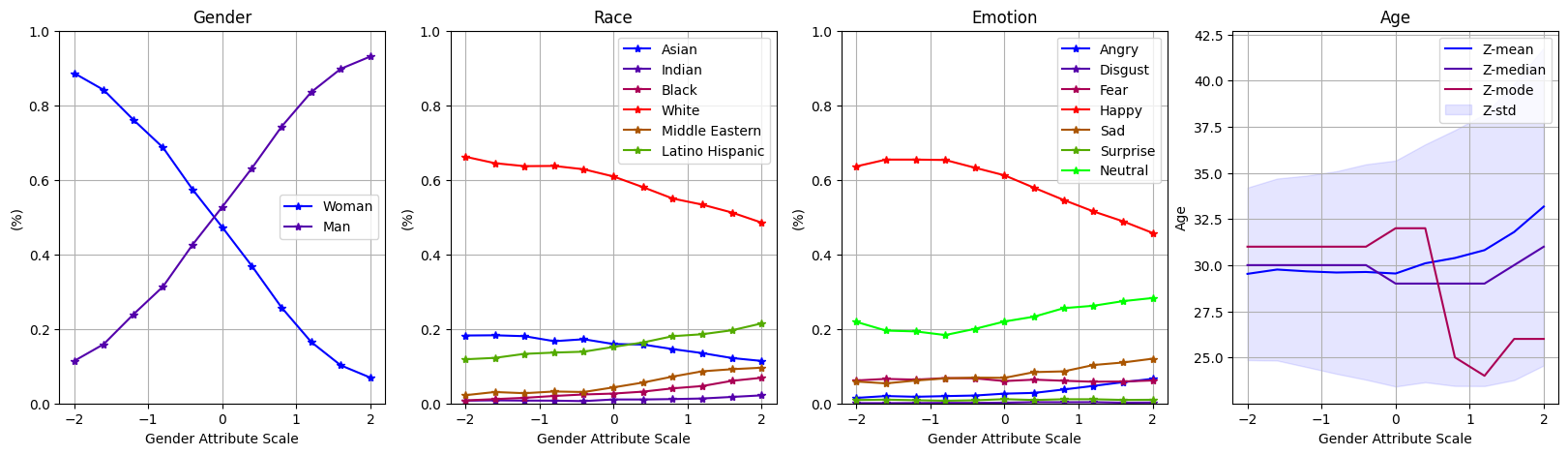}%
        \label{fig:a}%
    \end{subfigure}    
    \hfill
    \begin{subfigure}[b]{0.9\textwidth}
        \includegraphics[width=\textwidth]{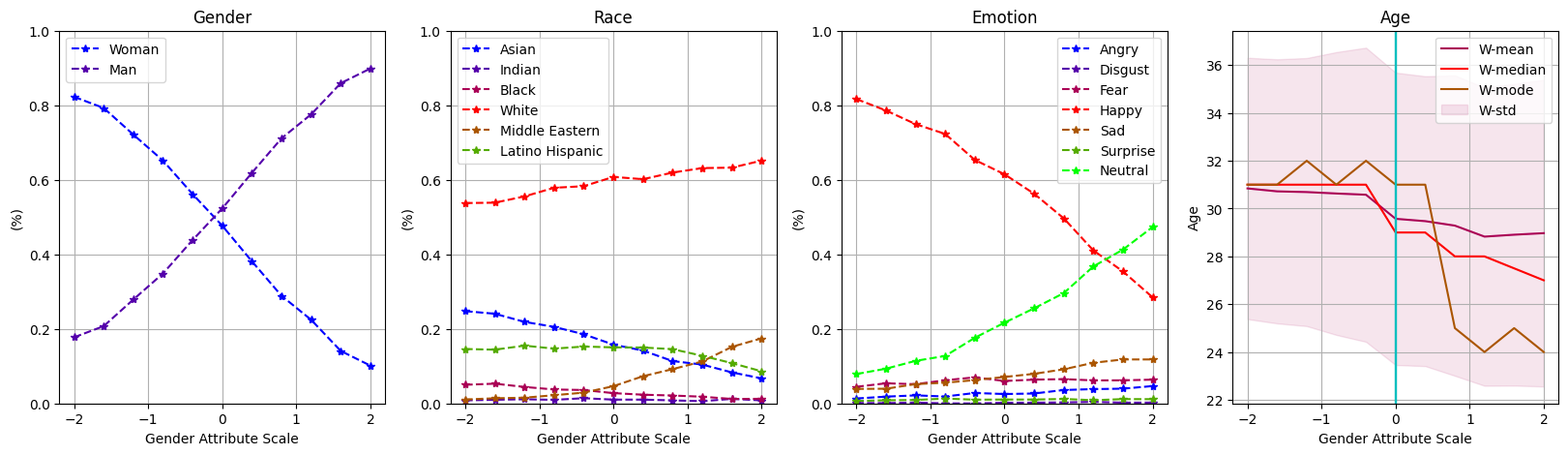}%
        \label{fig:b}%
    \end{subfigure}    
    \begin{subfigure}[b]{0.9\textwidth}
        \includegraphics[width=\textwidth]{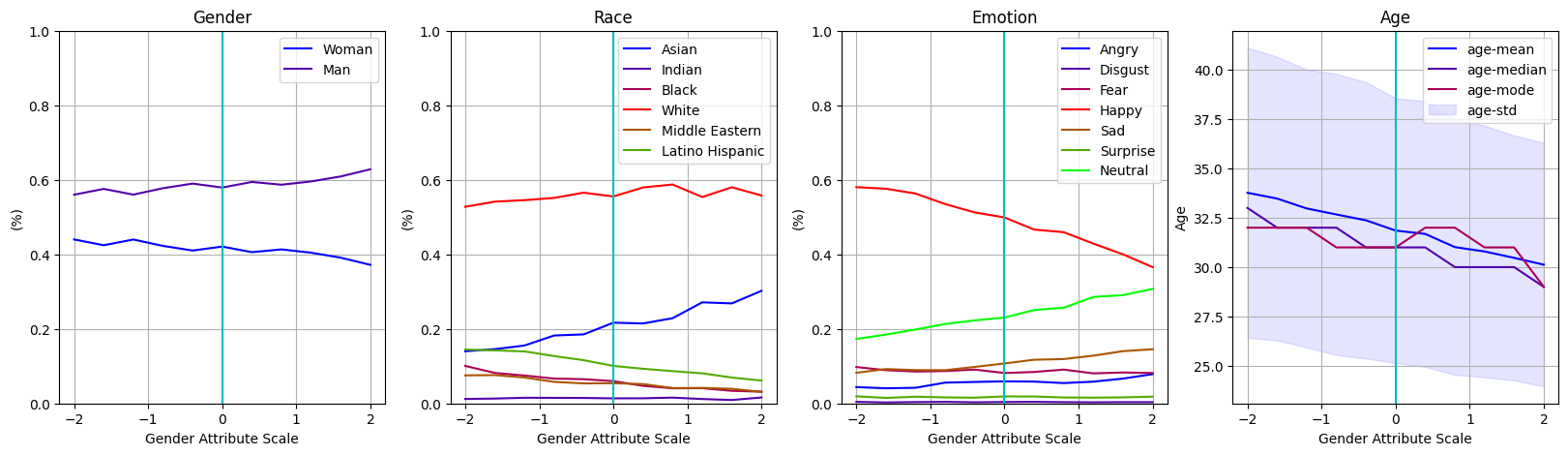}%
        \label{fig:c}%
    \end{subfigure}    
    \begin{subfigure}[b]{0.9\textwidth}
        \includegraphics[width=\textwidth]{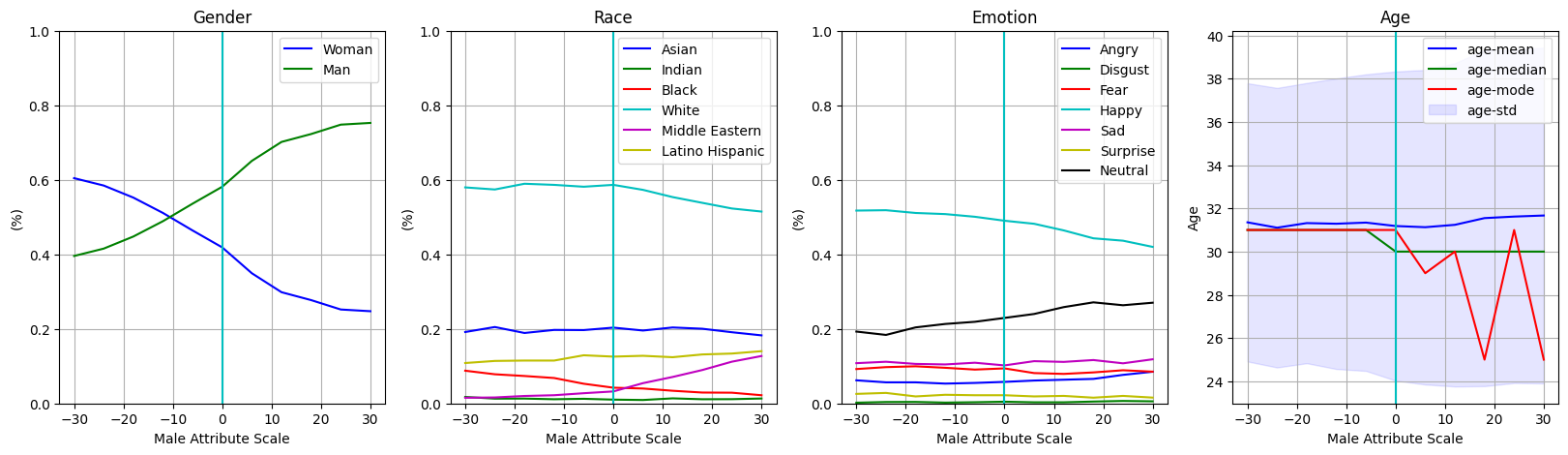}%
        \label{fig:d}%
    \end{subfigure}    
    \caption{Different embeddings spaces for each attribute controller: Z-space (row 1) and W-space (row 2) for interfacegan, PCA space (row 3 ) for GANSpace and style space (row 4) for stylespace algorithm.}\label{ctrl:gender_attr}
\end{figure}

\begin{figure}[H] 
    \centering
    \begin{subfigure}[b]{0.9\textwidth}
        \includegraphics[width=\textwidth]{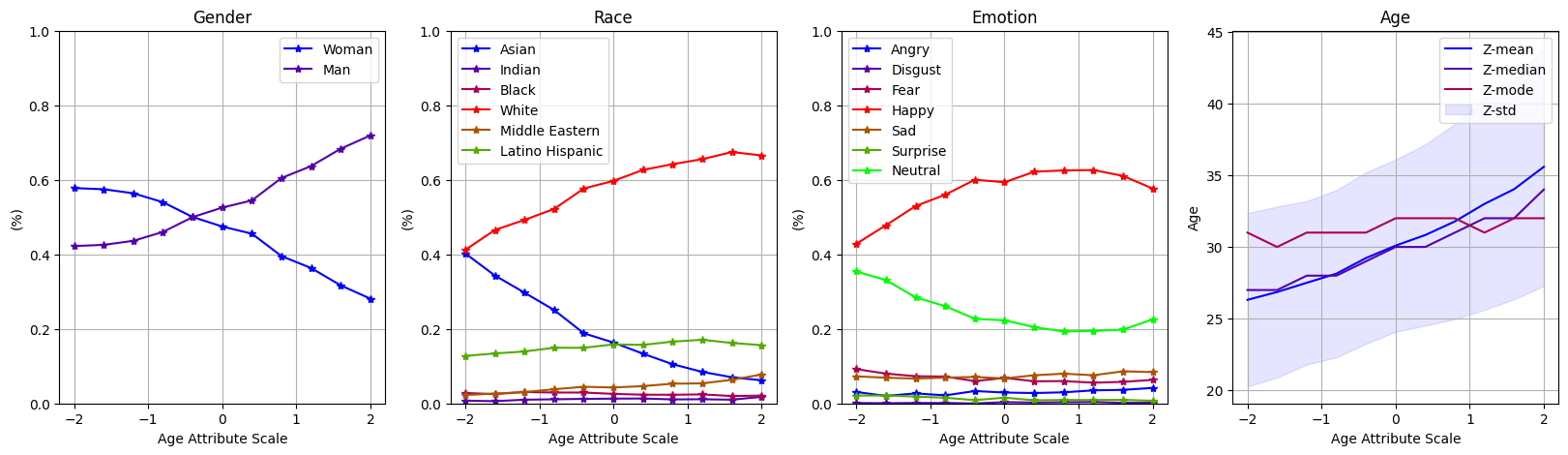}%
        \label{fig:a}%
    \end{subfigure}
    \begin{subfigure}[b]{0.9\textwidth}
        \includegraphics[width=\textwidth]{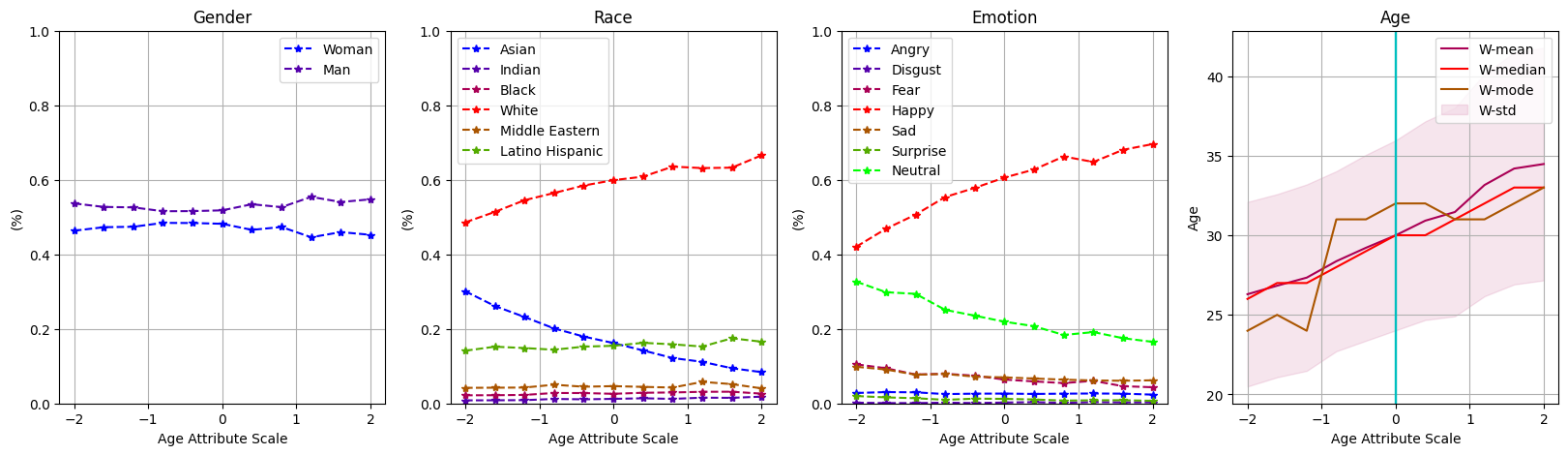}%
        \label{fig:b}%
    \end{subfigure}
    \begin{subfigure}[b]{0.9\textwidth}
        \includegraphics[width=\textwidth]{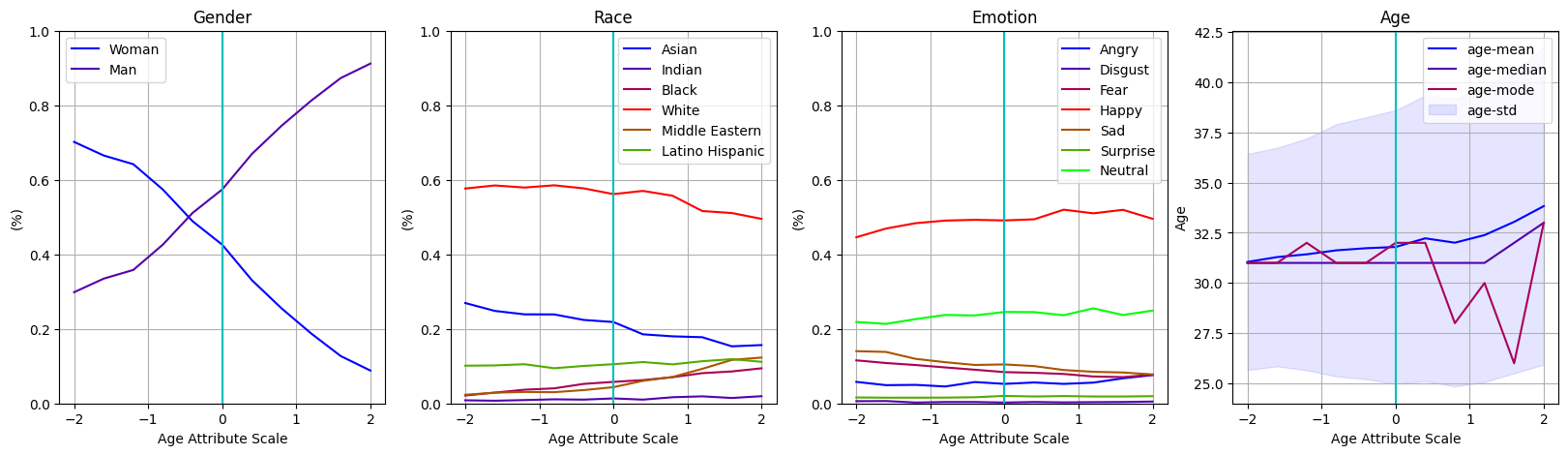}%
        \label{fig:c}%
    \end{subfigure}
    \begin{subfigure}[b]{0.9\textwidth}
        \includegraphics[width=\textwidth]{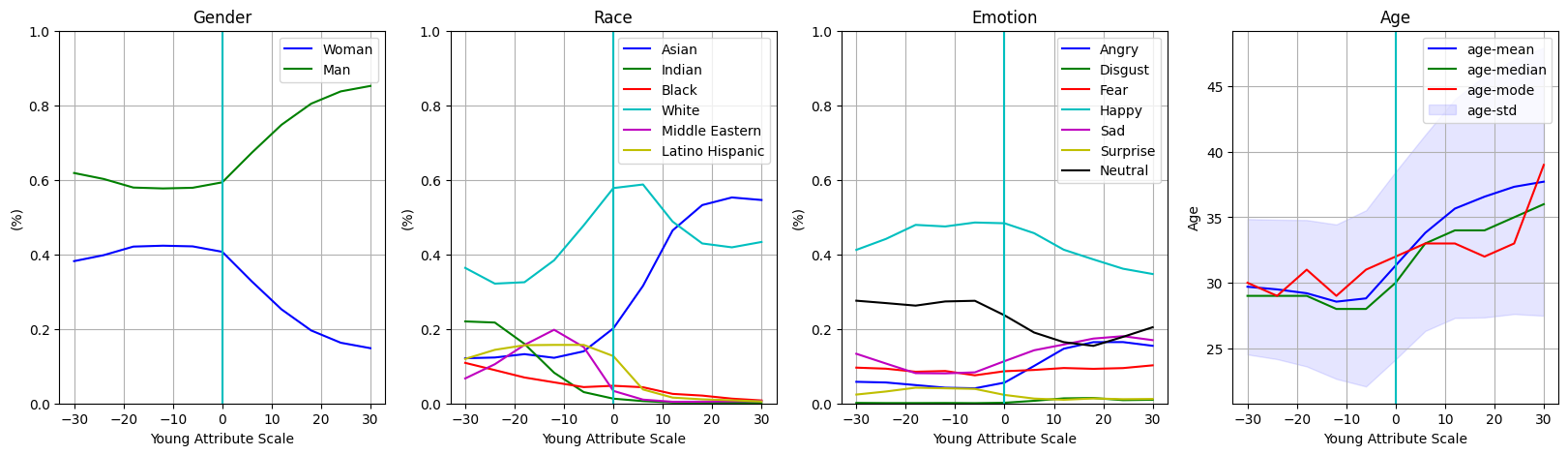}%
        \label{fig:d}%
    \end{subfigure}
    \caption{Image examples from: (a) CelebA. (b) Latent Diffusion Model. (c) StyleGAN2. (d) DDPM model}\label{ctrl:age_attr}
\end{figure}

\subsection{Two Channel Perturbation for Intergacegan}
\begin{figure}[H]
  \centering
  \includegraphics[width=\linewidth]{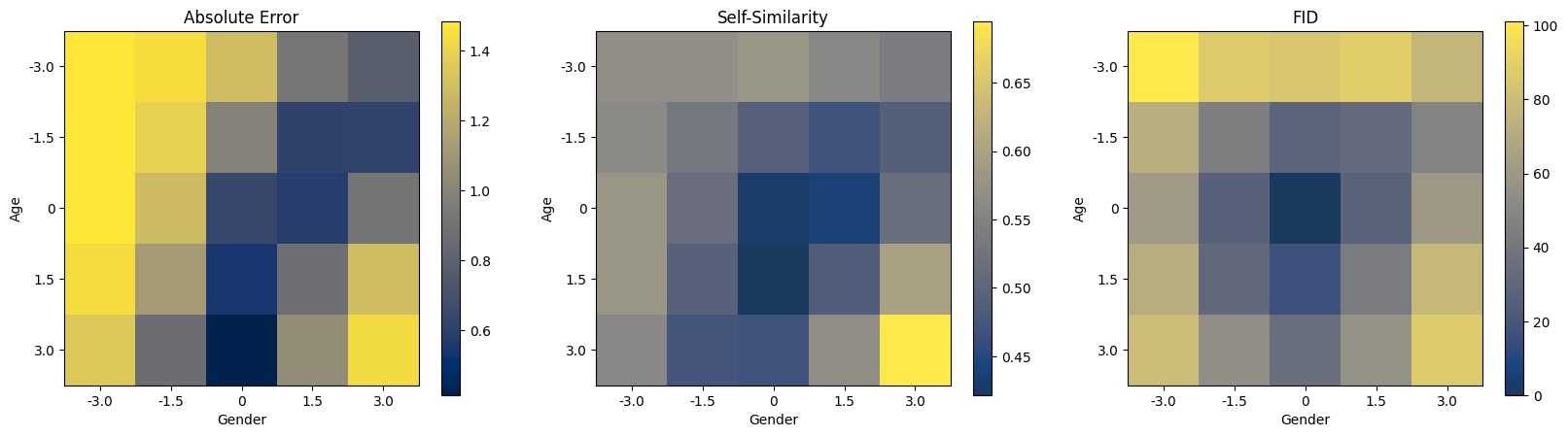}
  \caption{KL divergence, Sel-Similarity and FID score for two channel face attribute space.}\label{inter:2channel}
\end{figure}

\bibliographystyle{ACM-Reference-Format}
\bibliography{main}


\begin{thebibliography}{29}

%%% the Web address.

\ifx \showCODEN    \undefined \def \showCODEN     #1{\unskip}     \fi
\ifx \showDOI      \undefined \def \showDOI       #1{#1}\fi
\ifx \showISBNx    \undefined \def \showISBNx     #1{\unskip}     \fi
\ifx \showISBNxiii \undefined \def \showISBNxiii  #1{\unskip}     \fi
\ifx \showISSN     \undefined \def \showISSN      #1{\unskip}     \fi
\ifx \showLCCN     \undefined \def \showLCCN      #1{\unskip}     \fi
\ifx \shownote     \undefined \def \shownote      #1{#1}          \fi
\ifx \showarticletitle \undefined \def \showarticletitle #1{#1}   \fi
\ifx \showURL      \undefined \def \showURL       {\relax}        \fi
% invisible to TeX
\providecommand\bibfield[2]{#2}
\providecommand\bibinfo[2]{#2}
\providecommand\natexlab[1]{#1}
\providecommand\showeprint[2][]{arXiv:#2}

\bibitem[Abdal et~al\mbox{.}(2021)]%
        {10.1145/3447648}
\bibfield{author}{\bibinfo{person}{Rameen Abdal}, \bibinfo{person}{Peihao Zhu},
  \bibinfo{person}{Niloy~J. Mitra}, {and} \bibinfo{person}{Peter Wonka}.}
  \bibinfo{year}{2021}\natexlab{}.
\newblock \showarticletitle{StyleFlow: Attribute-Conditioned Exploration of
  StyleGAN-Generated Images Using Conditional Continuous Normalizing Flows}.
\newblock \bibinfo{journal}{\emph{ACM Trans. Graph.}} \bibinfo{volume}{40},
  \bibinfo{number}{3}, Article \bibinfo{articleno}{21} (\bibinfo{date}{May}
  \bibinfo{year}{2021}), \bibinfo{numpages}{21}~pages.
\newblock
\showISSN{0730-0301}
\urldef\tempurl%
\url{https://doi.org/10.1145/3447648}
\showDOI{\tempurl}


\bibitem[Anokhin et~al\mbox{.}(2020)]%
        {anokhin2020image}
\bibfield{author}{\bibinfo{person}{Ivan Anokhin}, \bibinfo{person}{Kirill
  Demochkin}, \bibinfo{person}{Taras Khakhulin}, \bibinfo{person}{Gleb
  Sterkin}, \bibinfo{person}{Victor Lempitsky}, {and} \bibinfo{person}{Denis
  Korzhenkov}.} \bibinfo{year}{2020}\natexlab{}.
\newblock \showarticletitle{Image Generators with Conditionally-Independent
  Pixel Synthesis}.
\newblock \bibinfo{journal}{\emph{arXiv preprint arXiv:2011.13775}}
  (\bibinfo{year}{2020}).
\newblock


\bibitem[Buolamwini and Gebru(2018)]%
        {pmlr-v81-buolamwini18a}
\bibfield{author}{\bibinfo{person}{Joy Buolamwini} {and}
  \bibinfo{person}{Timnit Gebru}.} \bibinfo{year}{2018}\natexlab{}.
\newblock \showarticletitle{Gender Shades: Intersectional Accuracy Disparities
  in Commercial Gender Classification}. In
  \bibinfo{booktitle}{\emph{Proceedings of the 1st Conference on Fairness,
  Accountability and Transparency}} \emph{(\bibinfo{series}{Proceedings of
  Machine Learning Research}, Vol.~\bibinfo{volume}{81})},
  \bibfield{editor}{\bibinfo{person}{Sorelle~A. Friedler} {and}
  \bibinfo{person}{Christo Wilson}} (Eds.). \bibinfo{publisher}{PMLR},
  \bibinfo{pages}{77--91}.
\newblock
\urldef\tempurl%
\url{https://proceedings.mlr.press/v81/buolamwini18a.html}
\showURL{%
\tempurl}


\bibitem[Choi et~al\mbox{.}(2020)]%
        {choi2020fair}
\bibfield{author}{\bibinfo{person}{Kristy Choi}, \bibinfo{person}{Aditya
  Grover}, \bibinfo{person}{Trisha Singh}, \bibinfo{person}{Rui Shu}, {and}
  \bibinfo{person}{Stefano Ermon}.} \bibinfo{year}{2020}\natexlab{}.
\newblock \showarticletitle{Fair generative modeling via weak supervision}. In
  \bibinfo{booktitle}{\emph{International Conference on Machine Learning}}.
  PMLR, \bibinfo{pages}{1887--1898}.
\newblock


\bibitem[Goodfellow et~al\mbox{.}(2016)]%
        {Goodfellow-et-al-2016}
\bibfield{author}{\bibinfo{person}{Ian Goodfellow}, \bibinfo{person}{Yoshua
  Bengio}, {and} \bibinfo{person}{Aaron Courville}.}
  \bibinfo{year}{2016}\natexlab{}.
\newblock \bibinfo{booktitle}{\emph{Deep Learning}}.
\newblock \bibinfo{publisher}{MIT Press}.
\newblock
\newblock
\shownote{\url{http://www.deeplearningbook.org}}.


\bibitem[Ho et~al\mbox{.}(2020)]%
        {ho2020denoising}
\bibfield{author}{\bibinfo{person}{Jonathan Ho}, \bibinfo{person}{Ajay Jain},
  {and} \bibinfo{person}{Pieter Abbeel}.} \bibinfo{year}{2020}\natexlab{}.
\newblock \showarticletitle{Denoising diffusion probabilistic models}.
\newblock \bibinfo{journal}{\emph{Advances in Neural Information Processing
  Systems}}  \bibinfo{volume}{33} (\bibinfo{year}{2020}),
  \bibinfo{pages}{6840--6851}.
\newblock


\bibitem[Härkönen et~al\mbox{.}(2020)]%
        {härkönen2020ganspace}
\bibfield{author}{\bibinfo{person}{Erik Härkönen}, \bibinfo{person}{Aaron
  Hertzmann}, \bibinfo{person}{Jaakko Lehtinen}, {and} \bibinfo{person}{Sylvain
  Paris}.} \bibinfo{year}{2020}\natexlab{}.
\newblock \showarticletitle{GANSpace: Discovering Interpretable GAN Controls}.
  In \bibinfo{booktitle}{\emph{Proc. NeurIPS}}.
\newblock


\bibitem[Kammoun et~al\mbox{.}(2022)]%
        {kammoun2022generative}
\bibfield{author}{\bibinfo{person}{Amina Kammoun}, \bibinfo{person}{Rim Slama},
  \bibinfo{person}{Hedi Tabia}, \bibinfo{person}{Tarek Ouni}, {and}
  \bibinfo{person}{Mohmed Abid}.} \bibinfo{year}{2022}\natexlab{}.
\newblock \showarticletitle{Generative Adversarial Networks for face
  generation: A survey}.
\newblock \bibinfo{journal}{\emph{Comput. Surveys}} \bibinfo{volume}{55},
  \bibinfo{number}{5} (\bibinfo{year}{2022}), \bibinfo{pages}{1--37}.
\newblock


\bibitem[Karakas et~al\mbox{.}(2022)]%
        {karakas2022fairstyle}
\bibfield{author}{\bibinfo{person}{Cemre Karakas}, \bibinfo{person}{Alara
  Dirik}, \bibinfo{person}{Eylul Yalcınkaya}, {and} \bibinfo{person}{Pinar
  Yanardag}.} \bibinfo{year}{2022}\natexlab{}.
\newblock \showarticletitle{FairStyle: Debiasing StyleGAN2 with Style Channel
  Manipulations}.
\newblock \bibinfo{journal}{\emph{ArXiv}}  \bibinfo{volume}{abs/2202.06240}
  (\bibinfo{year}{2022}).
\newblock


\bibitem[Karras et~al\mbox{.}(2019)]%
        {karras2019style}
\bibfield{author}{\bibinfo{person}{Tero Karras}, \bibinfo{person}{Samuli
  Laine}, {and} \bibinfo{person}{Timo Aila}.} \bibinfo{year}{2019}\natexlab{}.
\newblock \showarticletitle{A style-based generator architecture for generative
  adversarial networks}. In \bibinfo{booktitle}{\emph{Proceedings of the
  IEEE/CVF conference on computer vision and pattern recognition}}.
  \bibinfo{pages}{4401--4410}.
\newblock


\bibitem[Karras et~al\mbox{.}(2020)]%
        {karras2020analyzing}
\bibfield{author}{\bibinfo{person}{Tero Karras}, \bibinfo{person}{Samuli
  Laine}, \bibinfo{person}{Miika Aittala}, \bibinfo{person}{Janne Hellsten},
  \bibinfo{person}{Jaakko Lehtinen}, {and} \bibinfo{person}{Timo Aila}.}
  \bibinfo{year}{2020}\natexlab{}.
\newblock \showarticletitle{Analyzing and improving the image quality of
  stylegan}. In \bibinfo{booktitle}{\emph{Proceedings of the IEEE/CVF
  conference on computer vision and pattern recognition}}.
  \bibinfo{pages}{8110--8119}.
\newblock


\bibitem[Kawar et~al\mbox{.}(2022)]%
        {kawar2022imagic}
\bibfield{author}{\bibinfo{person}{Bahjat Kawar}, \bibinfo{person}{Shiran
  Zada}, \bibinfo{person}{Oran Lang}, \bibinfo{person}{Omer Tov},
  \bibinfo{person}{Huiwen Chang}, \bibinfo{person}{Tali Dekel},
  \bibinfo{person}{Inbar Mosseri}, {and} \bibinfo{person}{Michal Irani}.}
  \bibinfo{year}{2022}\natexlab{}.
\newblock \showarticletitle{Imagic: Text-based real image editing with
  diffusion models}.
\newblock \bibinfo{journal}{\emph{arXiv preprint arXiv:2210.09276}}
  (\bibinfo{year}{2022}).
\newblock


\bibitem[Krizhevsky et~al\mbox{.}(2009)]%
        {krizhevsky2009learning}
\bibfield{author}{\bibinfo{person}{Alex Krizhevsky}, \bibinfo{person}{Geoffrey
  Hinton}, {et~al\mbox{.}}} \bibinfo{year}{2009}\natexlab{}.
\newblock \showarticletitle{Learning multiple layers of features from tiny
  images}.
\newblock  (\bibinfo{year}{2009}).
\newblock


\bibitem[Lievrouw and Pope(1994)]%
        {lievrouw1994contemporary}
\bibfield{author}{\bibinfo{person}{Leah~A Lievrouw} {and}
  \bibinfo{person}{Janice~T Pope}.} \bibinfo{year}{1994}\natexlab{}.
\newblock \showarticletitle{Contemporary art as aesthetic innovation: Applying
  the diffusion model in the art world}.
\newblock \bibinfo{journal}{\emph{Knowledge}} \bibinfo{volume}{15},
  \bibinfo{number}{4} (\bibinfo{year}{1994}), \bibinfo{pages}{373--395}.
\newblock


\bibitem[McDuff et~al\mbox{.}(2019)]%
        {mcduff2019characterizing}
\bibfield{author}{\bibinfo{person}{Daniel McDuff}, \bibinfo{person}{Shuang Ma},
  \bibinfo{person}{Yale Song}, {and} \bibinfo{person}{Ashish Kapoor}.}
  \bibinfo{year}{2019}\natexlab{}.
\newblock \showarticletitle{Characterizing bias in classifiers using generative
  models}.
\newblock \bibinfo{journal}{\emph{Advances in neural information processing
  systems}}  \bibinfo{volume}{32} (\bibinfo{year}{2019}).
\newblock


\bibitem[Ning et~al\mbox{.}(2020)]%
        {ning2020multi}
\bibfield{author}{\bibinfo{person}{Xin Ning}, \bibinfo{person}{Fangzhe Nan},
  \bibinfo{person}{Shaohui Xu}, \bibinfo{person}{Lina Yu}, {and}
  \bibinfo{person}{Liping Zhang}.} \bibinfo{year}{2020}\natexlab{}.
\newblock \showarticletitle{Multi-view frontal face image generation: a
  survey}.
\newblock \bibinfo{journal}{\emph{Concurrency and Computation: Practice and
  Experience}} (\bibinfo{year}{2020}), \bibinfo{pages}{e6147}.
\newblock


\bibitem[Rombach et~al\mbox{.}(2021)]%
        {rombach2021highresolution}
\bibfield{author}{\bibinfo{person}{Robin Rombach}, \bibinfo{person}{Andreas
  Blattmann}, \bibinfo{person}{Dominik Lorenz}, \bibinfo{person}{Patrick
  Esser}, {and} \bibinfo{person}{Björn Ommer}.}
  \bibinfo{year}{2021}\natexlab{}.
\newblock \bibinfo{title}{High-Resolution Image Synthesis with Latent Diffusion
  Models}.
\newblock
\newblock
\showeprint[arxiv]{2112.10752}~[cs.CV]


\bibitem[Sattigeri et~al\mbox{.}(2019)]%
        {sattigeri2019fairness}
\bibfield{author}{\bibinfo{person}{Prasanna Sattigeri},
  \bibinfo{person}{Samuel~C Hoffman}, \bibinfo{person}{Vijil Chenthamarakshan},
  {and} \bibinfo{person}{Kush~R Varshney}.} \bibinfo{year}{2019}\natexlab{}.
\newblock \showarticletitle{Fairness GAN: Generating datasets with fairness
  properties using a generative adversarial network}.
\newblock \bibinfo{journal}{\emph{IBM Journal of Research and Development}}
  \bibinfo{volume}{63}, \bibinfo{number}{4/5} (\bibinfo{year}{2019}),
  \bibinfo{pages}{3--1}.
\newblock


\bibitem[Serengil and Ozpinar(2020)]%
        {serengil2020lightface}
\bibfield{author}{\bibinfo{person}{Sefik~Ilkin Serengil} {and}
  \bibinfo{person}{Alper Ozpinar}.} \bibinfo{year}{2020}\natexlab{}.
\newblock \showarticletitle{LightFace: A Hybrid Deep Face Recognition
  Framework}. In \bibinfo{booktitle}{\emph{2020 Innovations in Intelligent
  Systems and Applications Conference (ASYU)}}. IEEE, \bibinfo{pages}{23--27}.
\newblock
\urldef\tempurl%
\url{https://doi.org/10.1109/ASYU50717.2020.9259802}
\showDOI{\tempurl}


\bibitem[Serengil and Ozpinar(2021)]%
        {serengil2021lightface}
\bibfield{author}{\bibinfo{person}{Sefik~Ilkin Serengil} {and}
  \bibinfo{person}{Alper Ozpinar}.} \bibinfo{year}{2021}\natexlab{}.
\newblock \showarticletitle{HyperExtended LightFace: A Facial Attribute
  Analysis Framework}. In \bibinfo{booktitle}{\emph{2021 International
  Conference on Engineering and Emerging Technologies (ICEET)}}. IEEE,
  \bibinfo{pages}{1--4}.
\newblock
\urldef\tempurl%
\url{https://doi.org/10.1109/ICEET53442.2021.9659697}
\showDOI{\tempurl}


\bibitem[Shen et~al\mbox{.}(2020)]%
        {shen2020interfacegan}
\bibfield{author}{\bibinfo{person}{Yujun Shen}, \bibinfo{person}{Ceyuan Yang},
  \bibinfo{person}{Xiaoou Tang}, {and} \bibinfo{person}{Bolei Zhou}.}
  \bibinfo{year}{2020}\natexlab{}.
\newblock \showarticletitle{InterFaceGAN: Interpreting the Disentangled Face
  Representation Learned by GANs}.
\newblock \bibinfo{journal}{\emph{TPAMI}} (\bibinfo{year}{2020}).
\newblock


\bibitem[Song and Xiao(2015)]%
        {song2015lsun}
\bibfield{author}{\bibinfo{person}{Fisher Yu Yinda Zhang~Shuran Song} {and}
  \bibinfo{person}{Ari Seff~Jianxiong Xiao}.} \bibinfo{year}{2015}\natexlab{}.
\newblock \showarticletitle{LSUN: Construction of a Large-scale Image Dataset
  using Deep Learning with Humans in the Loop}.
\newblock \bibinfo{journal}{\emph{arXiv preprint arXiv:1506.03365}}
  (\bibinfo{year}{2015}).
\newblock


\bibitem[Tan et~al\mbox{.}(2020)]%
        {tan2020fairgen}
\bibfield{author}{\bibinfo{person}{Shuhan Tan}, \bibinfo{person}{Yujun Shen},
  {and} \bibinfo{person}{Bolei Zhou}.} \bibinfo{year}{2020}\natexlab{}.
\newblock \showarticletitle{Improving the Fairness of Deep Generative Models
  without Retraining}.
\newblock \bibinfo{journal}{\emph{arXiv preprint arXiv:2012.04842}}
  (\bibinfo{year}{2020}).
\newblock


\bibitem[Wang et~al\mbox{.}(2022)]%
        {wang2022gan}
\bibfield{author}{\bibinfo{person}{Xin Wang}, \bibinfo{person}{Hui Guo},
  \bibinfo{person}{Shu Hu}, \bibinfo{person}{Ming-Ching Chang}, {and}
  \bibinfo{person}{Siwei Lyu}.} \bibinfo{year}{2022}\natexlab{}.
\newblock \showarticletitle{Gan-generated faces detection: A survey and new
  perspectives}.
\newblock \bibinfo{journal}{\emph{arXiv preprint arXiv:2202.07145}}
  (\bibinfo{year}{2022}).
\newblock


\bibitem[Wolfe and Caliskan(2022)]%
        {wolfe2022markedness}
\bibfield{author}{\bibinfo{person}{Robert Wolfe} {and} \bibinfo{person}{Aylin
  Caliskan}.} \bibinfo{year}{2022}\natexlab{}.
\newblock \showarticletitle{Markedness in visual semantic ai}. In
  \bibinfo{booktitle}{\emph{2022 ACM Conference on Fairness, Accountability,
  and Transparency}}. \bibinfo{pages}{1269--1279}.
\newblock


\bibitem[Wu et~al\mbox{.}(2021)]%
        {wu2021stylespace}
\bibfield{author}{\bibinfo{person}{Zongze Wu}, \bibinfo{person}{Dani
  Lischinski}, {and} \bibinfo{person}{Eli Shechtman}.}
  \bibinfo{year}{2021}\natexlab{}.
\newblock \showarticletitle{Stylespace analysis: Disentangled controls for
  stylegan image generation}. In \bibinfo{booktitle}{\emph{Proceedings of the
  IEEE/CVF Conference on Computer Vision and Pattern Recognition}}.
  \bibinfo{pages}{12863--12872}.
\newblock


\bibitem[Xu et~al\mbox{.}(2018)]%
        {xu2018fairgan}
\bibfield{author}{\bibinfo{person}{Depeng Xu}, \bibinfo{person}{Shuhan Yuan},
  \bibinfo{person}{Lu Zhang}, {and} \bibinfo{person}{Xintao Wu}.}
  \bibinfo{year}{2018}\natexlab{}.
\newblock \showarticletitle{Fairgan: Fairness-aware generative adversarial
  networks}. In \bibinfo{booktitle}{\emph{2018 IEEE International Conference on
  Big Data (Big Data)}}. IEEE, \bibinfo{pages}{570--575}.
\newblock


\bibitem[Yang et~al\mbox{.}(2022)]%
        {yang2022diffusion}
\bibfield{author}{\bibinfo{person}{Ling Yang}, \bibinfo{person}{Zhilong Zhang},
  \bibinfo{person}{Yang Song}, \bibinfo{person}{Shenda Hong},
  \bibinfo{person}{Runsheng Xu}, \bibinfo{person}{Yue Zhao},
  \bibinfo{person}{Yingxia Shao}, \bibinfo{person}{Wentao Zhang},
  \bibinfo{person}{Bin Cui}, {and} \bibinfo{person}{Ming-Hsuan Yang}.}
  \bibinfo{year}{2022}\natexlab{}.
\newblock \showarticletitle{Diffusion models: A comprehensive survey of methods
  and applications}.
\newblock \bibinfo{journal}{\emph{arXiv preprint arXiv:2209.00796}}
  (\bibinfo{year}{2022}).
\newblock


\bibitem[Yu et~al\mbox{.}(2020)]%
        {yu2020inclusive}
\bibfield{author}{\bibinfo{person}{Ning Yu}, \bibinfo{person}{Ke Li},
  \bibinfo{person}{Peng Zhou}, \bibinfo{person}{Jitendra Malik},
  \bibinfo{person}{Larry Davis}, {and} \bibinfo{person}{Mario Fritz}.}
  \bibinfo{year}{2020}\natexlab{}.
\newblock \showarticletitle{Inclusive gan: Improving data and minority coverage
  in generative models}. In \bibinfo{booktitle}{\emph{Computer Vision--ECCV
  2020: 16th European Conference, Glasgow, UK, August 23--28, 2020,
  Proceedings, Part XXII 16}}. Springer, \bibinfo{pages}{377--393}.
\newblock


\end{thebibliography}

\end{document}